\documentclass[11pt]{article} 
\usepackage{tabularx}
\usepackage{threeparttable}
\usepackage{booktabs} 
\usepackage{amsfonts}
\usepackage{amsmath}
\usepackage{graphicx}
\usepackage{subcaption}
\usepackage{hyperref}
\usepackage{longtable}
\usepackage{makecell}
\usepackage[table]{xcolor}
\usepackage{lipsum}

\title{Autonomous mobile robot operations in production logistics: a dataset of jobs, dispatch events and robot states}
\author{Jan-Felix Klein$^{1,*}$, Yongkuk Jeong$^{1,2}$\\[6pt]
{\small $^{1}$Department of Production Engineering, KTH Royal Institute of Technology, Stockholm, Sweden}\\
{\small $^{2}$Digital Futures, KTH Royal Institute of Technology, Stockholm, Sweden}\\[6pt]
{\small $^{*}$Corresponding author: jfklein@kth.se}}
\date{}
\begin{document}
\maketitle
\begin{abstract} 
Autonomous mobile robots (AMRs) increasingly perform material transport in production logistics, where their operation is governed by job generation, dispatching and robot control. We present MoRoOp, a dataset of AMR operations recorded in a laboratory kit preparation and supply scenario over nine eight-hour shifts. During each shift, an AMR executed stochastically generated kit supply, empty-box refill and charging jobs. The dataset links job specifications, the operations constituting each job, dispatch events documenting operation state transitions and outcomes, and robot-state observations comprising position, orientation, velocity, per-wheel state of charge and diagnostics. It contains 1,382 jobs, 4,815 operations, 19,352 dispatch events and 140,386 robot-state observations together with the kit specifications used during job generation. The dataset was recorded in an operating laboratory environment, and technically valid observations of delays, obstructed navigation and unsuccessful operations were retained. Both raw and cleaned robot-state tables are provided. Documented reuse directions include the evaluation of AI agents on operational decision records, disturbance detection, operation prediction, data-driven simulation and event-log analysis.
\end{abstract}

\section*{Background \& Summary} 
Production logistics comprises the storage, handling and transport activities that supply and connect production processes. Increasing product variety, smaller batch sizes and more frequently reconfigured production structures raise the complexity of internal material flows and the effort required to manage them \cite{fragapane_increasing_2022}. Autonomous mobile robots (AMRs) have become a widely used resource in this context. In contrast to automated guided vehicles, which follow predetermined paths and are coordinated by a central controller, AMRs localize and plan their motion autonomously, and transport tasks can be assigned to them dynamically during execution \cite{fragapane_planning_2021, deryck_automated_2020}. The associated planning and control decisions, including task allocation, dispatching, routing and charging, have been studied extensively in intralogistics and warehousing research \cite{fragapane_planning_2021, azadeh_robotized_2019, pallottino_robotics_2026}.

Operating an AMR in such a system generates data at several levels. At the planning level, transport jobs are specified with attributes such as job type, requested content, release time and due date. At the execution level, a dispatching component decomposes jobs into executable operations, resolves their dependencies, and records state transitions, execution outcomes and termination reasons as discrete events. At the resource level, the AMR reports its position and orientation, velocity, navigation targets, battery state and diagnostic information as a time series. These levels differ in structure, temporal resolution and semantics, and they are typically produced and stored by different software components. They are consequently also analysed separately. Order and event data are treated in planning, control and process analysis, while robot telemetry is treated in navigation, diagnostics and maintenance. Making such heterogeneous sources jointly usable is a recurring requirement in production logistics \cite{tao_data-driven_2018, zhong_big_2015}.

The motivation for this dataset follows from that separation. A job record states what was requested and when it was due. The associated operation and dispatch-event records state how the request was decomposed and what happened to each part of it, including timeouts and unsuccessful executions. The robot-state observations describe the physical behaviour of the resource during the same period. When these records are available together, the executed behaviour of the system can be reconstructed and related to what was planned, for instance whether a late job followed from a delayed release, an extended waiting period at a station, a navigation disturbance or an intermediate charging decision. The same correspondence between plan and execution is the input required by digital shadows and digital twins of material handling systems \cite{kritzinger_digital_2018, pan_digital_2021} and by methods that construct simulation models directly from recorded operational data \cite{lugaresi_automated_2021}. A recent review of publicly available manufacturing datasets groups them into machine-level sensor recordings on one side and management-level production records on the other, and notes that documentation of the monitored processes is often incomplete \cite{golendukhina_review_2024}.

The dataset was recorded in the laboratory demonstrator associated with the concept of Self-Organizing Production Logistics (SOPL) \cite{klein_toward_2026}. SOPL addresses production logistics systems that are exposed to variability, dynamic interdependencies and frequent disturbances, and proposes multi-agent architectures in which embodied and non-embodied agents coordinate through event-driven interaction, shared semantic knowledge structures and digital twins. The order-driven kit preparation and supply scenario introduced in that work is the scenario used here, although no data from the present campaign were reported there. During the data-collection campaign, jobs were generated by fixed stochastic rules and executed under a rule-based dispatching policy rather than by agent-based decision-making. The released records therefore document the operation of the demonstrator under a rule-based control policy and provide a reference for later configurations of the same physical system.

Publicly available datasets involving mobile robots in shared indoor environments are largely oriented towards perception and navigation. Datasets such as TH\"OR-MAGNI provide human and robot trajectories recorded with motion capture and on-board sensors \cite{schreiter_thormagni_2025}, and comparable recordings have been published for a mobile robot performing retail tasks alongside human operators \cite{chen_human_2022}. A second group of datasets releases machine-level signals without the work that produced them, ranging from high-frequency signals of a stationary manipulator recorded per motion cycle \cite{brockmann_voraus_2024} to long sensor recordings of a single industrial asset annotated with reported failures \cite{veloso_metropt_2022}. Event logs published for process mining research provide case and activity structure with administrative timestamps, but contain no physical state of the executing resources \cite{zou_tvalp_2024}. Datasets combining both perspectives are less common. Examples include a multiphase flow facility benchmark that pairs an operation log with process measurements \cite{stief_pronto_2019} and a smart-factory dataset that pairs OPC UA logs with sensor measurements of stationary stations \cite{kaupp_context_2020}. In research on AMR dispatching and routing, evaluation is commonly based on generated benchmark instances \cite{meyer_beami_2024}.

The released dataset comprises nine eight-hour shifts recorded on five consecutive days and contains 1,382 jobs, 4,815 operations, 19,352 dispatch events and 140,386 robot-state observations. One further shift was recorded but excluded from the release because of an implementation error in the job generator, as documented in the Technical Validation section. Three characteristics of the released data are relevant for reuse. First, the data were recorded in an operating laboratory in which personnel could cross or temporarily obstruct the planned trajectory of the AMR. Waiting, delayed execution, obstructed navigation, operation timeouts and unsuccessful operations occurred during normal operation and were retained. The quality-screening procedure removed only technically invalid robot-state records, such as invalid localization states or physically implausible values, and the raw records are released alongside the cleaned ones. These records were retained deliberately, since they document the operating conditions under which the campaign took place. Second, the robot state was polled at a fixed frequency but stored only when a reported value changed, so the robot-state tables form an irregularly sampled time series. Third, station processing times were assigned by the job generator and represent waiting periods of the AMR rather than measurements of physical human work.

The recording documents the operation of a single AMR. This reflects the configuration of the demonstrator at the time of data collection and keeps the correspondence between dispatching decisions, executed operations and robot-state observations unambiguous, because every state observation belongs to the resource that executed the operation under consideration. Effects arising from the interaction of several vehicles, such as congestion, right-of-way conflicts and fleet-level task allocation, are consequently not represented. The demonstration roadmap defined for SOPL foresees increasingly distributed configurations \cite{klein_toward_2026}, for which the present recording provides a single-resource reference.

The combination of job specifications, operation records, dispatch events and robot-state observations from the same system allows several forms of reuse. Agents built on foundation models are increasingly proposed to plan and supervise the operation of physical logistics resources, which is also the direction taken by SOPL \cite{klein_toward_2026}. Assessing whether such an agent decides well requires a record of what was requested, of the executable plan that followed from the request, of what the resource physically did while carrying it out, and of what happened when execution was disturbed. Work on grounding language model plans in the actions a robot can perform \cite{ichter_do_2023}, on replanning from execution feedback \cite{huang_inner_2023} and on explaining failures from recorded robot experience \cite{liu_reflect_2023} rests on records of this kind, and applications of such models in manufacturing and supply chain operations are being surveyed \cite{zhang_survey_2025, song_large_2026}. In this dataset each job is present as the original request with its type, requested content and timing, as the ordered set of operations into which it was decomposed together with the dependencies between them, as the sequence of dispatcher events that queued, started and closed those operations, and as the robot-state record of the execution that followed, including the periods in which the AMR reported an obstruction. The decisions of an agent placed in the same situation can therefore be compared with the operations that the rule-based dispatcher produced and with the execution that resulted.

The robot-state tables constitute multivariate time series recorded during operation that included periods of obstructed navigation, and are therefore applicable to time-series anomaly detection and to fault detection and diagnosis for industrial robots \cite{darban_deep_2024, sabry_review_2024}. The reported stuck conditions, navigation states and per-wheel states of charge allow disturbances observed by the AMR itself to be located in time and, through the recorded timestamps, related to the operation that was being executed. The recorded release, start and finish timestamps, together with the due dates and execution outcomes of the jobs, support the prediction of operation durations and job lateness in analogy to predictive process monitoring on business process event logs \cite{ramamaneiro_deep_2023}.

Job and operation records also provide the arrival, routing and processing information required by methods that generate discrete-event simulation models and digital twins directly from operational data \cite{lugaresi_automated_2021, camargo_automated_2020, cimino_automatic_2025}, subject to the qualification noted above that station processing times originate from the job generator rather than from measured physical work. Dispatch events can be converted into event logs in which either jobs or operations serve as cases, which makes the data accessible to process mining techniques that have been applied to internal logistics event data in automotive production \cite{knoll_enabling_2019} and surveyed for Industry 4.0 settings more broadly \cite{akhramovich_systematic_2024}. Because jobs, operations, kits and stations are interrelated object types, the records can equally be prepared as object-centric event data. Finally, the scenario is small, physically interpretable and completely specified, which makes the dataset usable in teaching on production logistics, discrete-event simulation, process mining and industrial data analysis.

The dataset is released following the FAIR principles \cite{wilkinson_fair_2016}. Version~1.0.1 is archived in the KTH Data Repository \cite{klein_moroop_2026} and mirrored on Hugging Face \cite{klein_moroop_huggingface_2026}, and example code for loading and analysing the released tables is archived with the dataset toolkit \cite{klein_moroop_toolkit_2026}.


\section*{Methods}
\label{sec:methods}
The dataset was collected at the joint IPU laboratory facility of the Department of Production Engineering at KTH Royal Institute of Technology. Data acquisition was conducted over five consecutive days, comprising two eight-hour shifts per day.
During each shift, an AMR executed dynamically generated material-transport jobs within a laboratory production logistics scenario. The job generator created transport requests according to predefined stochastic rules, while the operation dispatcher translated these requests into executable operations and coordinated their execution on the AMR. The resulting dataset combines job specifications, operation records, dispatch events and robot-state observations. The following sections describe the experimental scenario, the procedure used to generate the jobs, and the data acquisition and processing pipeline.

\paragraph{Scenario description}
\begin{figure}[!ht]
\centering
\includegraphics[width=1.0\textwidth,keepaspectratio]{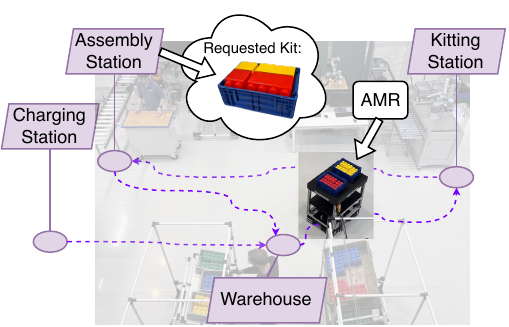}
\caption{An overview of the production logistics scenario at the IPU lab.}
\label{fig:scenario}
\end{figure}
The dataset was collected in a production logistics scenario previously introduced in \cite{klein_toward_2026}. The scenario comprises an AMR and four stations representing activities commonly encountered in production logistics environments, see Figure \ref{fig:scenario}. It implements a kit preparation and supply process in which an assembly station dynamically requests predefined kits.
Each kit is represented by a Eurobox containing a specified combination of components. In the laboratory implementation, the components are toy building bricks available in four colors and two sizes. Euroboxes are used as load carriers for both individual component supplies and completed kits. To distinguish these two uses, they are referred to as \textit{component boxes} and \textit{kit boxes}, respectively.
The principal assets of the scenario and their functions are as follows:
\begin{itemize}
    \item \textbf{Assembly Station}: The assembly station represents the internal customer of the production logistics process. It generates requests for specific kits to be prepared and delivered.
    \item \textbf{Warehouse}: The warehouse stores component boxes. Each component box contains building bricks of a specified color and provides the components required for kit preparation.
    \item \textbf{Kitting Station}: The kitting station represents the preparation of requested kits. At this station, the required bricks are transferred from the supplied component boxes into an empty kit box according to the corresponding kit specification.
    \item \textbf{AMR}: The AMR transports component and kit boxes between the stations. The AMR used for data collection was a wheel.me autonomous mobile robot (\href{https://www.wheel.me/en-us}{https://www.wheel.me/en-us}).
    \item \textbf{Charging Station}: The charging station provides the physical interface used by the AMR to recharge its battery.
\end{itemize}
A \textbf{kit supply job} consists of the following sequence of activities:
\begin{enumerate}
    \item The AMR navigates to the warehouse and waits while an operator loads the required component boxes.
    \item The AMR transports the component boxes to the kitting station and waits while the components are transferred into an empty kit box according to the corresponding kit specification.
    \item The AMR transports the completed kit box and the remaining partially filled component boxes to the assembly station and waits while the kit box is unloaded.
    \item The AMR returns the partially filled component boxes to the warehouse and waits while they are unloaded.
\end{enumerate}
In addition to \textbf{kit supply jobs}, the scenario includes two auxiliary job types: \textbf{empty-box refill jobs} and \textbf{charging jobs}. An empty-box refill job transports accumulated empty kit boxes from the assembly station to the kitting station. For this job type, the AMR navigates to the assembly station, waits while the empty boxes are loaded, transports them to the kitting station, and waits while they are unloaded. A charging job directs the AMR to the charging station, where it charges according to the charging policy described below.

\paragraph{Job generation}
\begin{figure}[!ht]
\centering
\includegraphics[width=0.55\textwidth,keepaspectratio]{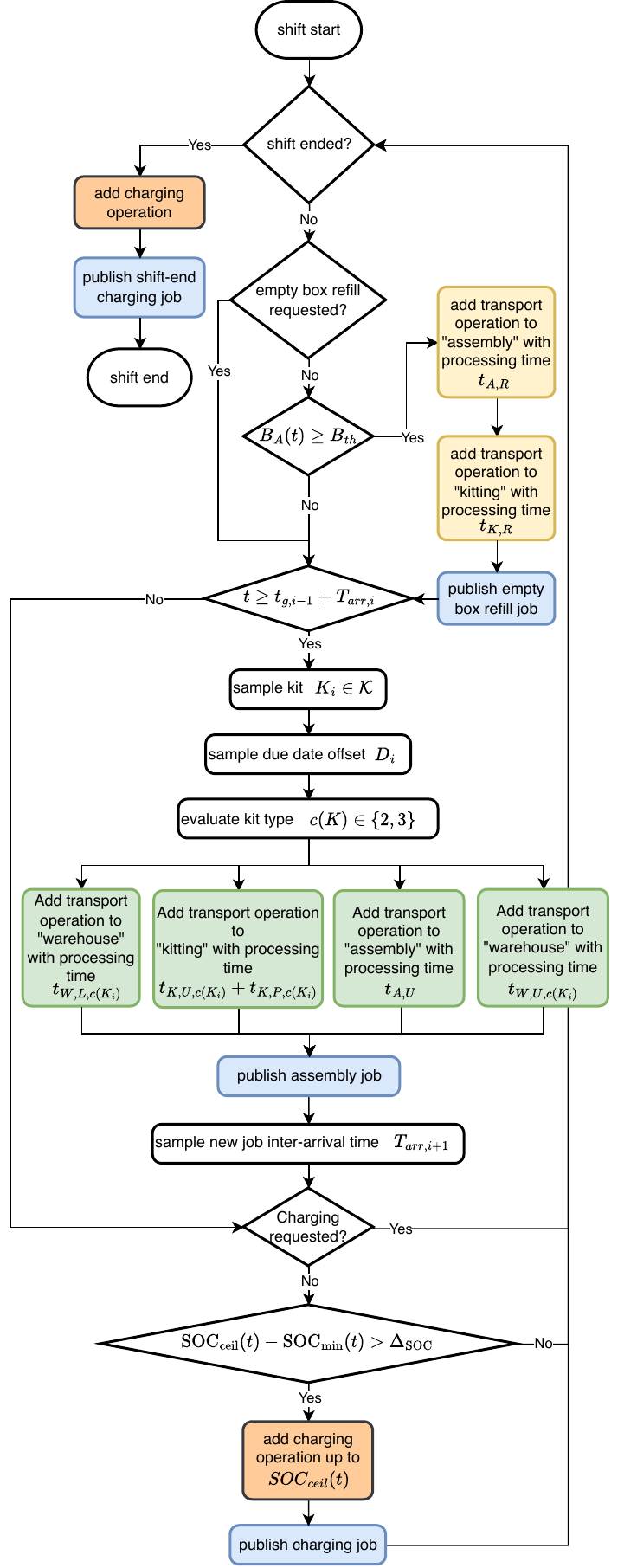}
\caption{Job generation procedure.}
\label{fig:flow_chart}
\end{figure}
During data collection, kit supply, empty-box refill, and charging jobs were generated by an event-driven job generator according to the workflow Fig.~\ref{fig:flow_chart}. 
At a periodic time interval, the generator evaluated the elapsed shift time, the scheduled generation time of the next kit supply job, the number of empty kit boxes accumulated at the assembly station, and the state of charge (SoC) of the AMR. The notation and parameter values used by the job generator are summarized in Table~\ref{tab:parameters}.

Kit supply jobs were indexed by $i$ and generated according to independently sampled inter-arrival times $T_{\mathrm{arr},i}$. Let $t$ denote the current time and $t_{g,i-1}$ the generation time of the preceding kit supply job. A new kit supply job $i$ was generated when

\begin{equation} 
t \geq t_{g,i-1} + T_{arr,i}. \label{eq:job_generation_condition}
\end{equation} 

After generation, the generation time was recorded as $t_{g,i}=t$, and the inter-arrival time for the subsequent job was sampled independently. The first kit supply job of each shift was generated at the start of the job-generation process.

For each kit supply job $i$, a kit $K_i$ was sampled uniformly from the set
$\mathcal{K}=\{K_{001},\ldots,K_{030}\}$.

Each kit specification defined a combination of building bricks and required components from two or three distinct colors, denoted by $c(K_i)\in\{2,3\}$. Because each component color was stored in a separate component box, $c(K_i)$ also determined whether two or three component boxes were required for the job.

A due-date offset $D_i$ was sampled independently and added to the job-generation time:
\begin{equation}
t_{d,i}=t_{g,i}+D_i, 
\end{equation}
where $t_{d,i}$ denotes the due date of kit supply job $i$.

Each kit supply job was instantiated as an ordered sequence of four operations corresponding to processing at the warehouse, kitting station, assembly station, and warehouse, respectively. The physical activities at these stations were represented by station-specific processing times. From the perspective of the AMR, each processing time corresponds to a waiting period at the associated station. Processing times were sampled independently from the triangular distributions specified in Table~\ref{tab:parameters}. For operations whose processing times depended on the number of component boxes, the corresponding distribution was selected according to $c(K_i)$. The waiting time assigned to the kitting operation was calculated as the sum of independently sampled component-unloading and kit-preparation times.

Empty-box refill jobs were generated based on the number of empty kit boxes accumulated at the assembly station, denoted by $B_A(t)$. The counter was incremented after the successful completion of each kit supply job. When

\begin{equation}
B_A(t) \geq B_{\mathrm{th}},
\end{equation} 
the generator created an empty-box refill job. The refill job comprised an operation at the assembly station for loading the accumulated empty kit boxes and a subsequent operation at the kitting station for unloading them. After successful completion of the refill job, $B_A(t)$ was reset to zero.

Intermediate charging jobs were considered during a shift. The AMR comprised four independently powered wheel units. The charging decision was based on the lowest state of charge among these units,
\begin{equation}
\mathrm{SOC}_{\min}(t) = \min_{r \in \mathcal{R}} \mathrm{SOC}_{r}(t).
\end{equation}
During a shift, the upper charging limit $\mathrm{SOC}_{\mathrm{ceil}}(t)$ decreased linearly from $75\%$ at the beginning of the shift to $40\%$ at its end:

\begin{equation}
\mathrm{SOC}_{\mathrm{ceil}}(t) = \mathrm{SOC}_{\mathrm{start}}
+\frac{t}{T_S}\left(\mathrm{SOC}_{\mathrm{end}}-\mathrm{SOC}_{\mathrm{start}}
\right).
\end{equation}

An intermediate charging job was generated when
\begin{equation}
\mathrm{SOC}_{\mathrm{ceil}}(t)-\mathrm{SOC}_{\min}(t)
\geq
\Delta_{\mathrm{SOC}},
\end{equation}

where $\Delta_{\mathrm{SOC}}$ was set to $5\%$. The time-dependent charging ceiling was used as the target SoC for the generated charging job.
At the end of each eight-hour shift, the generator stopped creating kit supply and empty-box refill jobs and appended a final charging job after the remaining operations. Unlike intermediate charging jobs, the final charging job did not specify a target SoC.

\newcolumntype{L}[1]{>{\raggedright\arraybackslash}p{#1}}
{
\scriptsize
\setlength{\tabcolsep}{4pt}
\renewcommand{\arraystretch}{1.05}
\setlength{\LTcapwidth}{\textwidth}

\begin{longtable}{@{}L{1.3cm} L{2.2cm} L{5.8cm} L{3cm}@{}}
\caption{Variables and parameters used in the job generator. All event times, durations, and processing times are specified in seconds. $\operatorname{Tri}(a,m,b)$ denotes a triangular distribution with minimum $a$, mode $m$, and maximum $b$. CRV denotes a continuous random variable; DRV denotes a discrete random variable, FP a fixed parameter.} 
\label{tab:parameters} \\
\toprule Symbol & Type & Description & Value or distribution \\ \midrule $i$ & Index & kit supply job index & $i \in \mathbb{N}$ \\ $t$ & Time variable & Current time relative to the beginning of the shift & $t \in [0,T_S]$ \\ $t_{g,i}$ & Event time & Generation time of kit supply job $i$ & $t_{g,i} \in [0,T_S]$ \\ $T_{arr,i}$ & CRV & Inter-arrival time preceding kit supply job $i$ & $\operatorname{Tri}(225,250,275)$ \\ $K_i$ & DRV & Kit sampled for kit supply job $i$ & $K_i \in \mathcal{K} = \{K_{001},\ldots,K_{030}\}$ \\ $c(K_i)$ & Derived variable & Number of unitized component boxes required for kit $K_i$ & $c(K_i) \in \{2,3\}$ \\ $D_i$ & CRV & Due-date offset assigned to kit supply job $i$ & $\operatorname{Tri}(275,285,295)$ \\ $B_A(t)$ & State variable & Number of empty boxes accumulated at the assembly station at time $t$ & $B_A(t) \in \mathbb{N}_0$ \\ $B_{\mathrm{th}}$ & FP & Empty-box threshold triggering a refill job & $3$ \\ 
$\mathcal{R}$ & Set & Set of individual wheel.me wheel units & $\mathcal{R}=\{BL,BR,TL,TR\}$ \\ $\mathrm{SOC}_{r}(t)$ & State variable & State of charge of wheel.me unit $r$ at time $t$ & $\mathrm{SOC}_{r}(t)\in[0,100]\,\%$, $r\in\mathcal{R}$ \\ 
$\mathrm{SOC}_{\min}(t)$ & Derived state variable & Lowest state of charge among the four wheel units & $\displaystyle \mathrm{SOC}_{\min}(t) =\min_{r\in\mathcal{R}}\mathrm{SOC}_{r}(t)$ \\ $\mathrm{SOC}_{\mathrm{ceil}}(t)$ & Time-dependent parameter & Upper charging ceiling at time $t$ & $\displaystyle \mathrm{SOC}_{\mathrm{start}} +\frac{t}{T_S} \left( \mathrm{SOC}_{\mathrm{end}} -\mathrm{SOC}_{\mathrm{start}} \right)$ \\ $\mathrm{SOC}_{\mathrm{start}}$ & FP & Upper charging ceiling at shift start & $75\,\%$ \\ $\mathrm{SOC}_{\mathrm{end}}$ & FP & Upper charging ceiling at shift end & $40\,\%$ \\ $\Delta_{\mathrm{SOC}}$ & FP & Minimum deficit below the charging ceiling required to trigger intermediate charging & $5\%$ \\
 $T_S$ & FP & Duration of a data-collection shift & $8\,\mathrm{h}$ \\ \midrule $t_{W,L,2}$ & CRV & Warehouse processing time for loading a two-box supply & $\operatorname{Tri}(5,7,10)$ \\ $t_{W,L,3}$ & CRV & Warehouse processing time for loading a three-box supply & $\operatorname{Tri}(7,10,15)$ \\ $t_{K,U,2}$ & CRV & Kitting processing time for unloading a two-box supply & $\operatorname{Tri}(5,7,10)$ \\ $t_{K,U,3}$ & CRV & Kitting processing time for unloading a three-box supply & $\operatorname{Tri}(7,10,15)$ \\ $t_{K,P,2}$ & CRV & Kit processing time for a two-box supply & $\operatorname{Tri}(10,12,15)$ \\ $t_{K,P,3}$ & CRV & Kit processing time for a three-box supply & $\operatorname{Tri}(15,20,25)$ \\ $t_{A,U}$ & CRV & Assembly processing time for unloading a completed kit & $\operatorname{Tri}(5,7,10)$ \\ $t_{W,U,2}$ & CRV & Warehouse processing time for unloading two returned boxes & $\operatorname{Tri}(5,7,10)$ \\ $t_{W,U,3}$ & CRV & Warehouse processing time for unloading three returned boxes & $\operatorname{Tri}(7,10,15)$ \\ $t_{A,R}$ & CRV & Assembly processing time for collecting empty boxes during a refill job & $\operatorname{Tri}(5,7,15)$ \\ $t_{K,R}$ & CRV & Kitting processing time for receiving empty boxes during a refill job & $\operatorname{Tri}(5,7,10)$ \\ \bottomrule
 \end{longtable}
 }

\paragraph{Data collection and processing pipeline}
\begin{figure}[!ht]
\centering
\includegraphics[width=1.0\textwidth,keepaspectratio]{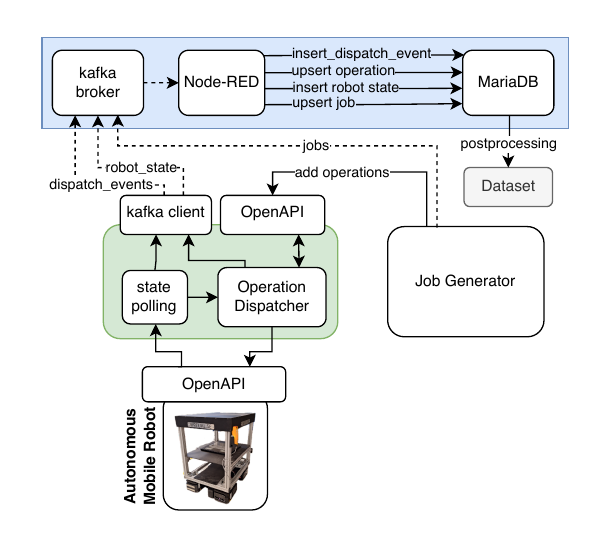}
\caption{The data collection pipeline.}
\label{fig:pipeline}
\end{figure}
The data collection pipeline connecting the AMR, job generator, operation dispatcher and data-storage infrastructure is illustrated in Fig.~\ref{fig:pipeline}. Apache Kafka was used for streaming data between the software components.

The AMR provides an OpenAPI interface for retrieving its current state and submitting navigation commands. During data collection, the AMR state was queried at a frequency of $1\,\mathrm{Hz}$. A new robot-state message was published to the \texttt{robot\_states} Kafka topic only when at least one reported state value differed from the previously retrieved state. Consequently, the robot state was evaluated at a fixed polling frequency, whereas the recorded observations form an event-triggered, irregularly sampled time series.

The operations instantiated by the job generator were managed by the \href{https://github.com/JaFeKl/OperationDispatcher}{OperationDispatcher} software module. The dispatcher maintained a queue of pending operations, evaluated operation dependencies, and coordinated their execution on the AMR. Information about generated jobs was published to the \texttt{jobs} topic, while operation state transitions and execution outcomes were published to the \texttt{dispatch\_events} topic. The resulting data collection pipeline therefore comprised three Kafka streams: \texttt{jobs}, \texttt{robot\_states}, and \texttt{dispatch\_events}.

Node-RED was used as the data-ingestion component. It consumed messages from the three Kafka streams, transformed the messages into SQL statements, and inserted the resulting records into a MariaDB database. Timestamps contained in the incoming messages were converted to ISO~8601 representation and standardized to Coordinated Universal Time (UTC) before storage.

Following completion of the data-collection campaign, the database records were exported and subjected to quality screening and curation. Quality screening of the robot-state data included identification and removal of records containing invalid localization states or physically implausible values. The original robot-state records were retained as the raw data record, while the screened records were stored separately as the cleaned robot-state table. No interpolation, resampling, smoothing, or imputation was applied during this process.

\section*{Data Records} 

Version~1.0.1 of the MoRoOp dataset is permanently archived in the KTH Data Repository \cite{klein_moroop_2026}. A copy of the same version is additionally available through Hugging Face \cite{klein_moroop_huggingface_2026} to facilitate programmatic access and dataset exploration. All tabular data are provided in Apache Parquet format. Parquet is a column-oriented file format that preserves the tabular schema and supports efficient storage and selective column access.

The dataset is organized around four related data entities: jobs, operations, dispatch events, and robot-state observations. Jobs represent material-transport requests generated by the job generator. Each job comprises one or more operations. Dispatch events record changes occurring during the lifecycle of individual operations. Robot-state observations provide time-stamped measurements and status information from the AMR. Jobs, operations and dispatch events can be related through their identifiers, while temporal alignment with robot-state observations is established using the recorded timestamps.

\begin{figure}[!b]
\centering
\includegraphics[width=0.8\textwidth,keepaspectratio]{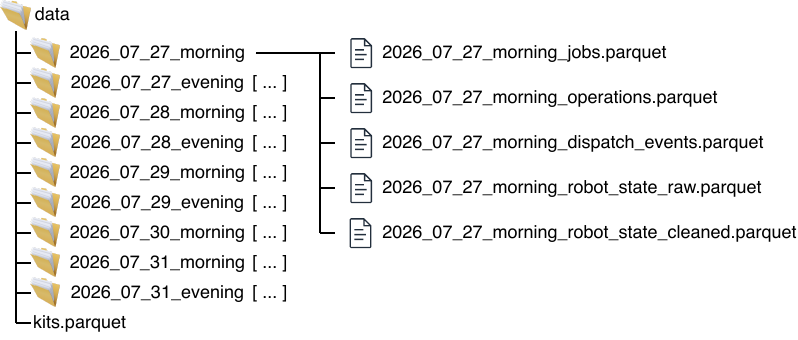}
\caption{The folder structure of the MoRoOp dataset.}
\label{fig:folder_structure}
\end{figure}

The organization of the dataset is illustrated in Fig.~\ref{fig:folder_structure}. The top-level \texttt{data} directory contains the dataset-level \texttt{kits.parquet} table and nine shift-specific subdirectories. The subdirectories follow the naming convention \texttt{YYYY\_MM\_DD\_\allowbreak shift}, where \texttt{YYYY\_MM\_DD} denotes the date of data collection and \texttt{shift} identifies the \texttt{morning} or \texttt{evening} shift. Each shift directory contains five Parquet tables: \texttt{dispatch\_events}, \texttt{jobs}, \texttt{operations}, \texttt{robot\_state\_raw}, and \texttt{robot\_state\_cleaned}. The file names include the corresponding date and shift identifier to preserve their association when individual files are accessed independently of the directory structure.

\paragraph{Kit definitions} The top-level \texttt{kits.parquet} defines the composition of the kits available to the job generator. The table uses a long-format representation in which each row specifies one component variant required for a kit. A component variant is defined by its color and size, together with the required quantity. Consequently, multiple rows with the same kit identifier collectively define the complete specification of one kit.

\begin{table}[!ht]
\centering
\caption{Fields contained in the \texttt{kits.parquet} table.}
\label{tab:kits_fields}
\footnotesize
\begin{tabularx}{\textwidth}{@{}l l X@{}}
\toprule Field & Data type & Description \\
\midrule
\texttt{kit\_id} & string & Unique identifier of the kit to which the brick variant belongs. Multiple rows may share the same kit identifier. \\
\texttt{color} & string & Color of the brick, either \texttt{blue}, \texttt{red}, \texttt{yellow} or \texttt{green}. \\ \texttt{size} & string & Size of the brick variant, either \texttt{small} or \texttt{large}. \\ \texttt{quantity} & int & Number of bricks of the specified type required by the kit. \\ \bottomrule
\end{tabularx}
\end{table}

\paragraph{Jobs} Each shift-specific job table contains the final recorded representation of the jobs created during the corresponding data-collection shift. Each row represents one job and is uniquely identified by the \texttt{id} field. The tables include the three job types described in the Methods section. The \texttt{operations} field contains a JSON-encoded ordered list of the operation identifiers constituting the job. These identifiers link each job to the corresponding records in the \texttt{*\_operations.parquet} and \texttt{*\_dispatch\_\allowbreak events.parquet} tables. The individual fields are defined in Table~\ref{tab:jobs_fields}.

\begin{table}[!ht]
\centering
\caption{Fields contained in the \texttt{*\_jobs.parquet} table.}
\label{tab:jobs_fields}
\footnotesize
\begin{tabularx}{\textwidth}{@{}l l X@{}}
\toprule Field & Data type & Description \\
\midrule
\texttt{id} & string & UUIDv4 unique identifier for the job. \\
\texttt{created\_at} & timestamp & Timestamp at which the job record was created. \\
\texttt{payload} & string & JSON-encoded job specification, including job type and kit identifier. \\
\texttt{operations} & string & JSON-encoded ordered list of operation identifiers belonging to the job. \\
\texttt{state} & string & Recorded lifecycle state of the job. \\
\texttt{outcome} & string & Recorded execution outcome of the job. \\
\texttt{termination\_reason} & string & Recorded reason for job termination. \\
\texttt{release\_date} & timestamp & Timestamp at which the job was released. \\
\texttt{start\_time} & timestamp  & Timestamp at which the execution started. \\
\texttt{due\_date} & timestamp & Scheduled due timestamp for the job. \\
\texttt{finish\_time} & timestamp & Timestamp at which the execution finished. \\
\bottomrule
\end{tabularx}
\end{table}

\paragraph{Operations}
The shift-specific \texttt{*\_operations.parquet} tables contain the final recorded representation of the operations instantiated during the corresponding data-collection shift. Each row represents one operation and is uniquely identified by the \texttt{id} field. The \texttt{job\_id} field links the operation to its parent job in the corresponding \texttt{*\_jobs.parquet} table. As each job comprises one or more operations, multiple operation records may reference the same job identifier. The \texttt{payload} field contains the operation specification, including the AMR goal location. The \texttt{dependencies} field specifies the dependencies on preceding operations used by the operation dispatcher to ensure the required execution order. The individual fields are defined in Table~\ref{tab:operations_fields}.

\begin{table}[!ht]
\centering
\caption{Fields contained in the \texttt{*\_operations.parquet} table.}
\label{tab:operations_fields}
\footnotesize
\begin{tabularx}{\textwidth}{@{}l l X@{}}
\toprule Field & Data type & Description \\
\midrule
\texttt{id} & string & UUIDv4 unique identifier for the operation. \\
\texttt{created\_at} & timestamp & Timestamp at which the operation record was created. \\
\texttt{payload} & string & JSON-encoded operation specification, including applicable operation-specific parameters. \\
\texttt{priority} & int & Numeric priority assigned to the operation. \\
\texttt{job\_id} & string & Identifier of the job to which the operation belongs. \\
\texttt{dependencies} & string & JSON-encoded list of dependencies linking the operation to required preceding operations. \\
\texttt{state} & string & Recorded lifecycle state of the operation. \\
\texttt{outcome} & string & Recorded execution outcome of the operation. \\
\texttt{termination\_reason} & string & Recorded reason for operation termination. \\
\texttt{retry\_count} & int & Number of retries recorded for the operation. \\
\texttt{start\_time} & timestamp  & Timestamp at which the execution started. \\
\texttt{finish\_time} & timestamp & Timestamp at which the execution finished. \\
\bottomrule
\end{tabularx}
\end{table}

\paragraph{Dispatch Events}
The shift-specific \texttt{*\_dispatch\_events.parquet} tables contain the events published by the \href{https://github.com/JaFeKl/OperationDispatcher}{OperationDispatcher} module during the corresponding data-collection shift. In contrast to the operations tables, which contain one final record for each operation, the dispatch-event tables document changes that occurred during operation execution. Multiple dispatch events may therefore refer to the same operation. Each event is uniquely identified by the \texttt{id} field. Where an event is associated with a specific operation, the \texttt{operation\_id} field links the event to the corresponding record in the \texttt{*\_operations.parquet} table. The \texttt{event\_type} field identifies the type of dispatcher event, while the JSON-encoded \texttt{changes} field describes the operation attributes affected by the event. Additional event-specific information may be included in the \texttt{meta\_data} field. The individual fields are defined in Table~\ref{tab:dispatch-events-fields}.

\begin{table}[!ht]
\centering
\caption{Fields contained in the \texttt{*\_dispatch\_events.parquet} table.}
\label{tab:dispatch-events-fields}
\footnotesize
\begin{tabularx}{\textwidth}{@{}l l X@{}}
\toprule
Field & Data type & Description \\
\midrule
\texttt{id} & string & UUIDv4 unique identifier for the dispatch event. \\
\texttt{created\_at} & timestamp & Timestamp at which the event was created. \\
\texttt{event\_type} & string & Recorded event type. \\
\texttt{operation\_id} & string & Identifier of the operation associated with the event, where applicable. \\
\texttt{changes} & string & JSON-encoded description of the operation attributes changed by the event. \\
\texttt{meta\_data} & string & JSON-encoded supplementary information associated with the event, where applicable. \\

\bottomrule
\end{tabularx}
\end{table}

\paragraph{Robot States}
Each shift directory contains a raw and a cleaned robot-state table. The \texttt{*\_robot\_state\_raw.parquet} table contains the robot-state and diagnostic data acquired from the AMR during the corresponding shift. The AMR state was polled at \(1\,\mathrm{Hz}\), but a record was published only when a change relative to the previously retrieved state was detected. Consequently, the robot-state tables do not constitute regularly sampled time series, and the interval between successive records may exceed one second.

The fields \texttt{current\_\allowbreak position\_\allowbreak id}, \texttt{navigating\_\allowbreak to\_\allowbreak position\_\allowbreak id}, and \texttt{last\_\allowbreak navigated\_\allowbreak position\_\allowbreak id} use the internal position identifiers of the AMR navigation system. Position identifiers 135, 136, 137, and 138 correspond to the charging station, warehouse, kitting station, and assembly station, respectively.

The \texttt{*\_robot\_state\_cleaned.parquet} table contains the corresponding robot-state records after the quality-screening procedure described in the Technical Validation section. Records containing invalid localization states or physically implausible values were excluded from the cleaned table. No interpolation, resampling, smoothing, or imputation was applied. The raw and cleaned robot-state tables follow the schema defined in Table~\ref{tab:robot-state-fields}.

\begin{table}[!ht]
\centering
\caption{Data dictionary for the \texttt{robot\_state} table.}
\label{tab:robot-state-fields}
\scriptsize
\begin{tabularx}{\textwidth}{@{}l l X@{}}
\toprule
Field & Data type & Description \\
\midrule
\texttt{id} & int & Sequential identifier assigned to observation. \\
\texttt{created\_at} & timestamp & Timestamp at which the robot state was recorded. \\
\texttt{operating\_mode} & string & Robot operating mode at the time of observation. \\
\texttt{state} & string & Robot state at the time of observation. \\
\texttt{stuck\_reason} & string & Reported reason the robot was stuck, if any. \\
\texttt{pos\_x} & double & Robot x-coordinate in the map reference frame. \\
\texttt{pos\_y} & double & Robot y-coordinate in the map reference frame. \\
\texttt{pos\_theta} & double & Robot orientation angle in the map reference frame. \\
\texttt{velocity\_x} & double & Robot linear velocity along the x-axis. \\
\texttt{velocity\_y} & double & Robot linear velocity along the y-axis. \\
\texttt{velocity\_omega} & double & Robot angular velocity. \\
\texttt{current\_position\_id} & int & Identifier of the robot's current named position, when available. \\
\texttt{navigating\_to\_position\_id} & int & Identifier of the navigation destination, when available. \\
\texttt{last\_navigated\_position\_id} & int & Identifier of the most recently navigated named position. \\
\texttt{next\_goal\_timestamp} & timestamp & Timestamp associated with the next navigation goal (robot clock). \\
\texttt{tl\_charge\_state} & int & SoC of front left wheel, expressed in $\%$ \\
\texttt{tr\_charge\_state} & int & SoC of front right wheel, expressed in $\%$ \\
\texttt{bl\_charge\_state} & int & SoC of rear left wheel, expressed in $\%$ \\
\texttt{br\_charge\_state} & int & SoC of rear right wheel, expressed in $\%$ \\
\texttt{is\_robot\_charging} & bool & Overall charge indicator. \\
\texttt{tl\_charging} & bool & Charge indicator for the front left wheel. \\
\texttt{tr\_charging} & bool & Charge indicator for the front right wheel. \\
\texttt{bl\_charging} & bool & Charge indicator for the rear left wheel. \\
\texttt{br\_charging} & bool & Charge indicator for the rear right wheel. \\
\texttt{robot\_uptime} & double & Robot uptime reported at the time of observation. \\
\texttt{diagnostics} & string & JSON-encoded robot diagnostic status messages. \\
\bottomrule
\end{tabularx}
\end{table}

\section*{Data Overview}
\begin{figure}[htbp]
    \centering
    
    \begin{subfigure}[b]{1.0\textwidth}
        \centering
        \includegraphics[width=\linewidth]{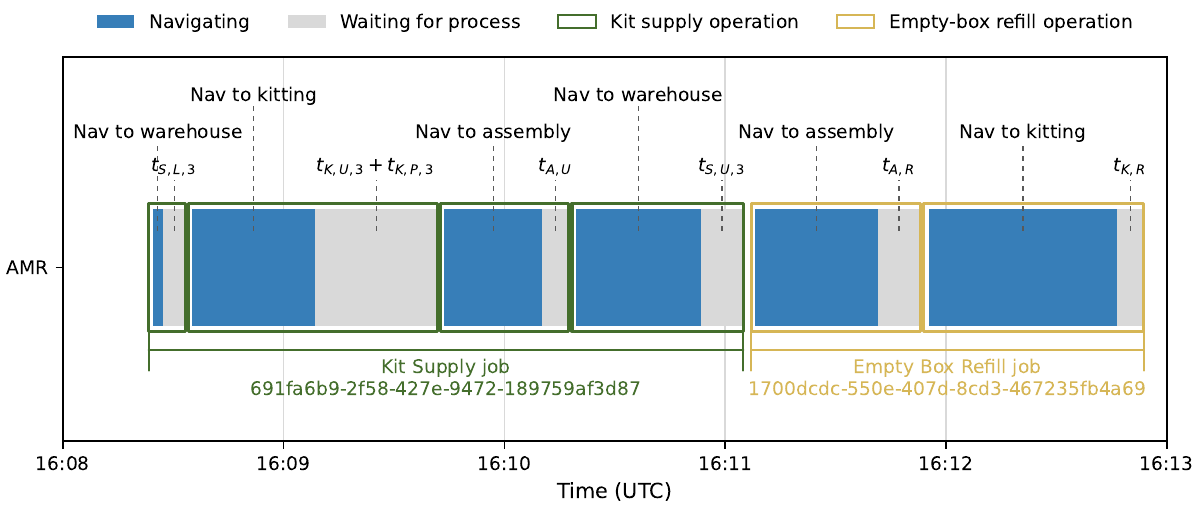}
        \caption{Gantt chart of the operations associated with a selected kit supply job and the subsequent empty-box refill job.}
        \label{fig:gantt}
    \end{subfigure}
    
    \vspace{0.3cm} 
    
    \begin{subfigure}[b]{1.0\textwidth}
        \centering
        \includegraphics[width=\linewidth]{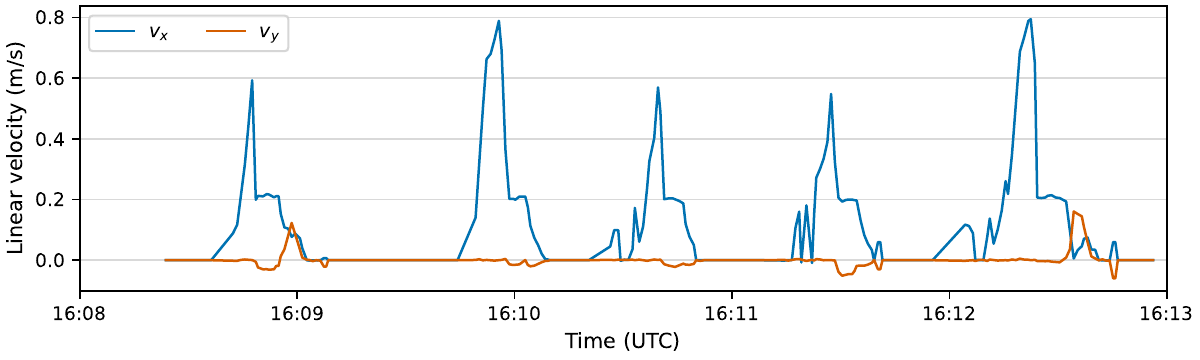}
        \caption{AMR velocity recorded during the corresponding operation sequence.}
        \label{fig:velocity}
    \end{subfigure}
    
    \caption{Representative operation sequence comprising the supply of kit \texttt{K014} and an empty-box refill job during the \texttt{2026\_07\_27\_evening} shift.}
    \label{fig:operations}
\end{figure}

\begin{figure}[htbp]
    \centering
    
    \begin{subfigure}[b]{1.0\textwidth}
        \centering
        \includegraphics[width=\linewidth]{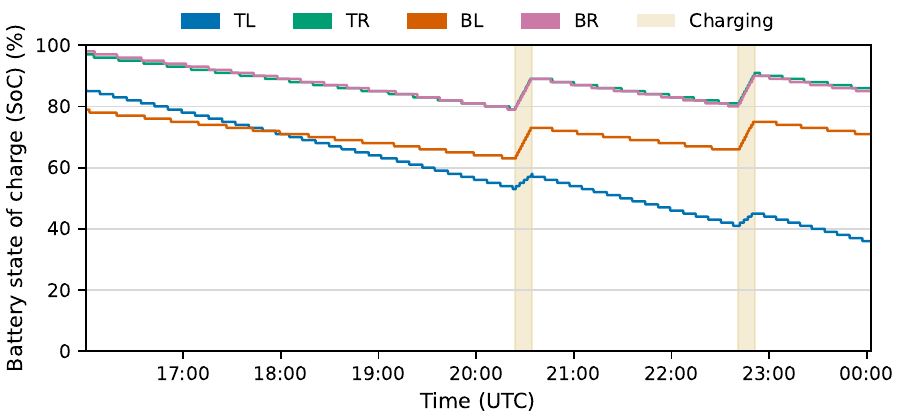}
        \caption{Individual wheel-unit states of charge and intermediate charging activity.}
        \label{fig:top_image}
    \end{subfigure}
    
    \vspace{0.3cm} 
    
    \begin{subfigure}[b]{1.0\textwidth}
        \centering
        \includegraphics[width=\linewidth]{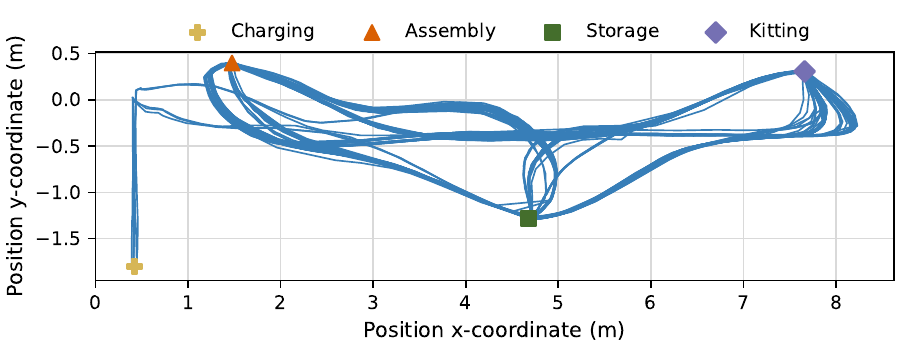}
        \caption{AMR trajectory derived from the cleaned robot-state data.}
        \label{fig:bottom_image}
    \end{subfigure}
    
    \caption{Representative robot-state data from the \texttt{2026\_07\_27\_evening} shift.}
    \label{fig:robot_state_overview}
\end{figure}

Figures~\ref{fig:operations} and \ref{fig:robot_state_overview} illustrate representative data contained in the dataset. Both figures use records from the \texttt{2026\_07\_27\_evening} shift. Figure~\ref{fig:operations} illustrates the temporal relationship between operation records and robot-state observations for a kit supply job followed by an empty-box refill job. Figure~\ref{fig:robot_state_overview} illustrates the types of battery and localization information available in the robot-state tables. The figures are intended to demonstrate how the released tables can be combined to reconstruct operational sequences and robot trajectories.

\section*{Technical Validation}
The experimental system and data collection pipeline were monitored throughout each shift using dedicated dashboards for the AMR and the job-generation infrastructure. This continuous monitoring allowed technical problems affecting the AMR, job execution, or data acquisition to be identified during operation.

Following data collection, the completeness and consistency of the recorded tables were evaluated separately for each shift. Table~\ref{tab:dataset-overview} reports the numbers of records contained in the respective shift-specific tables. Among the nine shifts included in the released dataset, the number of recorded jobs ranged from 148 to 156 per shift. The corresponding numbers of operations and dispatch events were also comparable across these shifts. These record counts indicate that the stochastic job-generation process produced broadly comparable workloads while retaining the intended variability in job inter-arrival times, sampled kits, sampled processing times and charging activities.

The \texttt{2026\_07\_30\_\allowbreak evening} shift initially contained considerably fewer records than the remaining shifts. Subsequent investigation identified an implementation error in the job-generation procedure. After an operation timeout occurred while the AMR was traveling to execute an empty-box refill job, the internal refill-job state was not reset correctly. Consequently, no further empty-box refill jobs were generated during the shift, although empty kit boxes continued to accumulate at the assembly station. The resulting recording therefore did not represent the job-generation procedure specified for the data collection campaign. The entire shift was excluded from the released dataset because the observed inconsistency resulted from a software implementation error rather than from a valid operational disturbance.

The internal consistency of the remaining records was evaluated using the identifiers and lifecycle information contained in the different data tables. Each job was checked for a valid unique identifier and references to its constituent operations. The operation identifiers were subsequently used to relate the jobs to the corresponding operation and dispatch-event records. Timestamps were checked for chronological consistency, and the recorded operation states and outcomes were examined for consistency with the operation lifecycle. These checks verified that the relationships among jobs, operations, and dispatch events were preserved in the released dataset.

The robot-state records underwent an additional quality-screening procedure. Records containing invalid localization states or physically implausible values were identified and removed. The screening criteria were restricted to technical validity and did not target unusual but valid operational behaviour. No interpolation, resampling, smoothing, or imputation was applied after the removal of invalid records. The raw robot-state tables are provided alongside the cleaned tables to make the curation process transparent and to allow users to apply alternative quality-screening procedures. The validation procedure deliberately retained technically valid observations of operational disturbances. Data collection was conducted in an operating laboratory environment in which personnel could cross or temporarily obstruct the planned trajectory of the AMR. Such interactions could result in waiting, changes in navigation behaviour, delayed operation execution, or unsuccessful operations. These records were retained when the associated robot states and system events were technically valid. The dataset therefore represents the operational conditions encountered during the data collection campaign rather than an idealized, disturbance-free execution of the production logistics scenario.

\begin{table}[!ht] \centering \caption{Number of records in each dataset table by shift.} \label{tab:dataset-overview} 
\scriptsize \setlength{\tabcolsep}{4pt} \renewcommand{\arraystretch}{1.1} 
\begin{tabular}{@{}lrrrrr@{}} 
\toprule 
\makecell[tl]{shift} & \makecell[tl]{dispatch\\events} & \makecell[tl]{jobs} & \makecell[tl]{operations} & \makecell[tl]{robot state\\(raw)} & \makecell[tl]{robot state\\(cleaned)} \\ 
\midrule 2026\_07\_27\_morning & 2,149 & 154 & 536 & 15,549 & 15,194 \\ 2026\_07\_27\_evening & 2,145 & 154 & 535 & 15,389 & 15,038 \\ 2026\_07\_28\_morning & 2,173 & 156 & 540 & 15,912 & 15,553 \\ 2026\_07\_28\_evening & 2,159 & 155 & 538 & 15,566 & 15,252 \\ 2026\_07\_29\_morning & 2,170 & 155 & 541 & 15,546 & 15,182 \\ 2026\_07\_29\_evening & 2,167 & 155 & 539 & 15,491 & 15,168 \\ 2026\_07\_30\_morning & 2,103 & 148 & 520 & 15,053 & 14,711 \\ \rowcolor{gray!20} 2026\_07\_30\_evening* & 1,873 & 121 & 468 & 12,217 & -- \\ 2026\_07\_31\_morning & 2,149 & 152 & 533 & 15,949 & 15,635 \\ 2026\_07\_31\_evening & 2,137 & 153 & 533 & 15,931 & 15,694 \\ \bottomrule 
\end{tabular}
\begin{tablenotes}
\footnotesize
\item *Shift excluded because of the job-generation error described in the Technical Validation section. 
\end{tablenotes}
\end{table}

\section*{Usage Notes}
All timestamps are expressed in ISO~8601 format and standardized to Coordinated Universal Time (UTC). Records from the different data streams can therefore be aligned temporally using their timestamps. Job, operation, and dispatch-event identifiers provide additional relational links between generated jobs and their execution history.

The robot-state data should be treated as an irregularly sampled time series. Although the AMR state was queried at $1\,\mathrm{Hz}$, a record was stored only when the reported state differed from the preceding observation. The time interval between successive robot-state observations can therefore exceed one second. Users requiring regularly sampled data should perform resampling as part of their analysis and select an interpolation or aggregation strategy appropriate for their application.

Both raw and cleaned robot-state tables are provided. The raw tables preserve the observations recorded by the data-ingestion pipeline, including observations subsequently identified as technically invalid. The cleaned tables exclude observations identified during the quality-screening procedure described in the Technical Validation section. Users developing alternative quality criteria are therefore encouraged to use the raw tables as the starting point for their own curation.

The relationships between jobs, operations, and dispatch events can be reconstructed using the corresponding identifiers. The temporal relationship between these records and robot-state observations can be established using their UTC timestamps. When reconstructing operation trajectories, users should account for the fact that station processing times represent waiting periods assigned to the AMR and are not direct measurements of physical human or automated processing times.

The dataset contains observations from a laboratory production-logistics environment. Consequently, the recorded robot behaviour reflects the specific physical layout, navigation configuration, job-generation parameters, AMR hardware, and software components used during the data-collection campaign. The dataset should therefore not be interpreted as a statistically representative sample of industrial AMR operations.

\section*{Data Availability} 



Version~1.0.1 of the MoRoOp dataset is permanently archived in the KTH Data Repository \cite{klein_moroop_2026} and constitutes the version of record associated with this publication. The repository contains the complete released dataset, including the kit definitions and the nine released shift-specific datasets.

A copy of Version 1.0.1 is also available through Hugging Face \cite{klein_moroop_huggingface_2026} to facilitate programmatic access and dataset exploration. The KTH Data Repository should be used as the archival source of record.

\section*{Code Availability} 
Example code for loading, accessing, and analyzing the MoRoOp dataset is available in the public \href{https://github.com/JaFeKl/moroop-dataset-toolkit} {MoRoOp Dataset Toolkit} repository. The repository also contains the scripts used to reproduce the figures presented in this publication. The code version associated with this publication is identified by release \texttt{v1.0.0}, an archived copy of which is deposited on Zenodo \cite{klein_moroop_toolkit_2026}.



\section*{Author Contributions}
J.-F.K. designed the study, collected the data, developed processing scripts, and wrote the manuscript.
Y.J. designed the study, wrote and reviewed the manuscript.

\section*{Funding}
Open Access funding provided by KTH Royal Institute of Technology. This work was supported by XPRES (Excellence in Production Research).

\section*{Competing Interests}
The authors declare no competing interests.

\end{document}